\documentclass[10pt,twocolumn,letterpaper]{article}

\usepackage{cvpr}              
\usepackage{float}

\usepackage{listings}
\usepackage{booktabs}
\usepackage{multirow}
\usepackage{threeparttable}
\usepackage{siunitx}
\usepackage{dsfont}
\usepackage{overpic}
\usepackage{contour}
\contourlength{1pt}
\usepackage{xspace}
\usepackage{sidecap}
\usepackage{rotating}
\usepackage[normalem]{ulem}
\usepackage{amsmath}
\usepackage{xcolor,colortbl}
\usepackage[dvipsnames]{xcolor}
\usepackage{tabularx}
\usepackage{soul}
\usepackage{url}
\usepackage{amsfonts}
\usepackage{comment}
\usepackage{booktabs}
\usepackage{etoolbox}

\newlength{\indentlaenge}
\definecolor{cvprblue}{rgb}{0.21,0.49,0.74}
\usepackage[pagebackref,breaklinks,colorlinks,allcolors=cvprblue]{hyperref}

\def\paperID{} 
\def\confName{CVPR}
\def\confYear{2026}

\title{Do Large Language Models Understand Data Visualization Principles?}
\author{
    \textbf{Martín Sinnona*}$^{1,3}$ \hspace{0.1cm} 
    \textbf{Valentín Bonás*}$^{1}$ \hspace{0.1cm}
    \textbf{Viviana Siless}$^{1}$ \hspace{0.1cm} 
    \textbf{Emmanuel Iarussi}$^{1,2}$ \hspace{0.1cm} \\
    Contributed equally $^{*}$ \\
    Universidad Torcuato Di Tella, Buenos Aires, Argentina $^{1}$\\
    Consejo Nacional de Investigaciones Científicas y Técnicas, Buenos Aires, Argentina $^{2}$ \\
    Universidad de Buenos Aires, Buenos Aires, Argentina $^{3}$ \\
}

\begin{document}
\maketitle

\begin{abstract}
Data visualization principles, derived from decades of research in design and perception, ensure proper visual communication.
While prior work has shown that large language models (LLMs) can generate charts or flag misleading figures, it remains unclear whether they—and their vision-language counterparts (VLMs)—can reason about and enforce visualization principles directly.
Constraint-based systems encode these principles as logical rules for precise automated checks, but translating them into formal specifications demands expert knowledge.
This motivates leveraging LLMs and VLMs as principle checkers that can reason about visual design directly, bypassing the need for symbolic rule specification.
In this paper, we present the first systematic evaluation of both LLMs and VLMs on their ability to reason about visualization principles, using hard-verification ground truth derived from Answer Set Programming (ASP).
We compiled a set of visualization principles expressed as natural-language statements and generated a controlled dataset of approximately 2,000 Vega-Lite specifications annotated with explicit principle violations, complemented by over 300 real-world Vega-Lite charts.
We evaluated both checking and fixing tasks, assessing how well models detect principle violations and correct flawed chart specifications. 
Our work highlights both the promise of large (vision-)language models as flexible validators and editors of visualization designs and the persistent gap with symbolic solvers on more nuanced aspects of visual perception.
They also reveal an interesting asymmetry: frontier models tend to be more effective at correcting violations than at detecting them reliably.
\end{abstract}    

\section{Introduction}

Clear visualization is essential for data communication. 
Decades of research in human perception, graphical integrity, and design best practices have been distilled into actionable principles that help ensure charts convey information accurately and without distortion \cite{tufte1983visual,cleveland1984graphical,huff1954lie,tufte1997visual,munzner2014visualization,cairo2019charts,ware2019information,healy2024data}.  
Authoring tools and validation frameworks have sought to operationalize these principles, enabling users to avoid misleading choices.
\begin{figure}[t]
    \centering
    \includegraphics[width=1\linewidth]{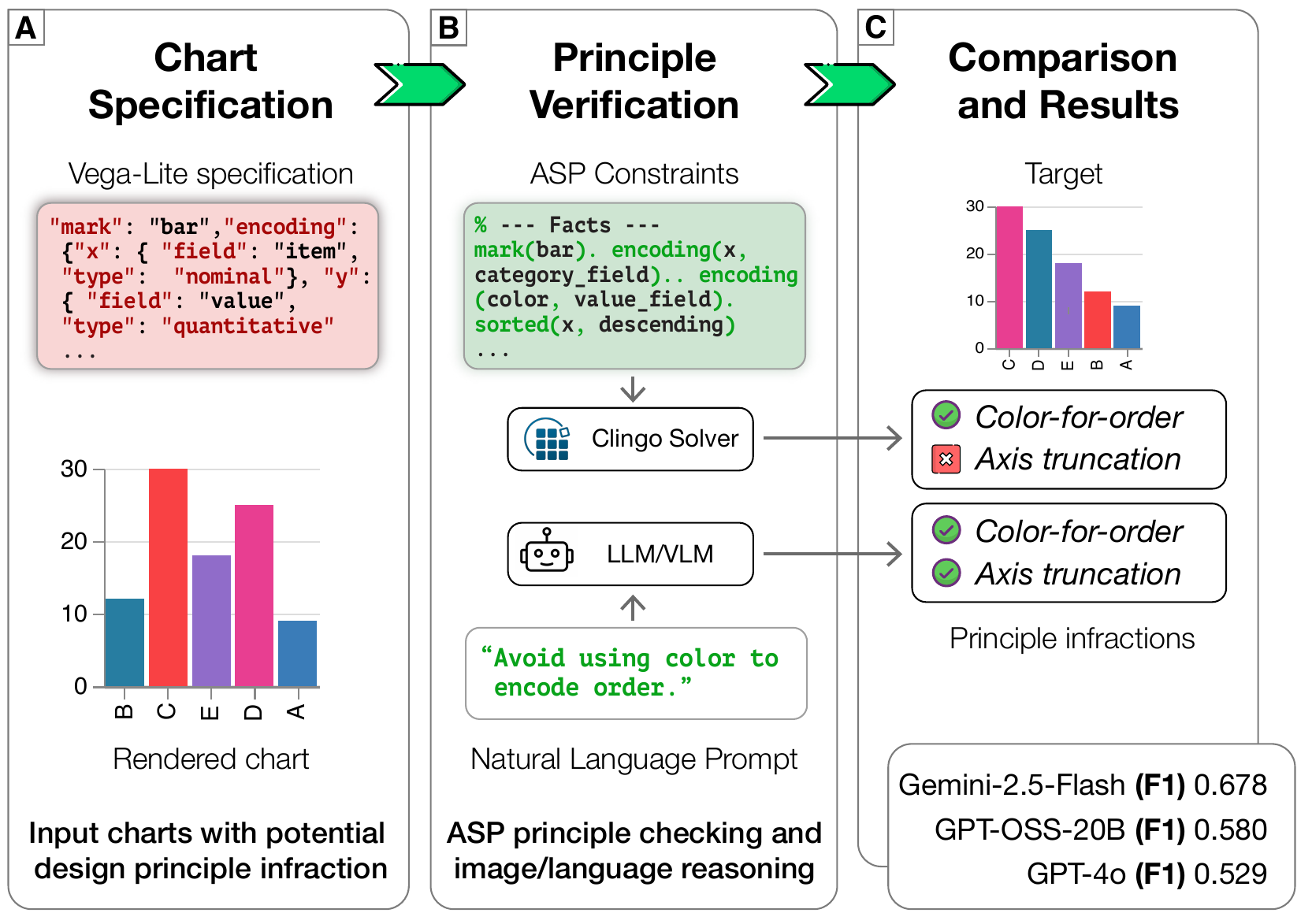}
    
    \caption{\textbf{Bridging symbolic and language-based reasoning for visualization principle checking.} Given a chart specification and its corresponding rendering (\textbf{A}), we automatically check compliance with visualization principles using two complementary reasoning approaches: formal verification via handcrafted ASP constraints and language/image-based reasoning from natural-language prompts and chart renderings (\textbf{B}). The detected infractions are then compared (\textbf{C}), with performance metrics evaluated across both checking and fixing tasks.}    
    \label{fig:figure_1}
\end{figure}
Constraint-based systems formalize visual guidelines as logical constraints, enabling automatic checks over chart specifications. 
Similarly, linter frameworks~\cite{chen2021vizlinter} carry these ideas into production by flagging and repairing violations in Vega-Lite~\cite{satyanarayan2016vega} specs. 
These approaches demonstrate that principles can be encoded precisely and enforced automatically. 
However, maintaining such principle sets requires specialized expertise and tooling (e.g., ASP with \emph{Clingo}~\cite{kaminski2017clingo}), limiting their scalability and flexibility.

Recent work suggests that large (vision-)language models may provide a promising alternative.
Lo et al.~\cite{lo2024good} demonstrate that multimodal LLMs, when carefully prompted, can identify a wide range of misleading chart patterns directly from images across twenty-one issue categories. 
Extending beyond detection, recent studies~\cite{hong2025llms,pandey2025benchmarking} examine whether LLMs exhibit \emph{visualization literacy}—the capacity to read, interpret, and reason about visual representations—using adaptations of the Visualization Literacy Assessment Test (VLAT)~\cite{lee2016vlat}.
Authors report that, although LLMs sometimes produce correct answers, their performance trails human baselines and frequently depends on prior knowledge instead of cues present in the visualization.
These studies explore complementary skills: one detecting misleading visual patterns, the others probing visualization understanding through question answering—highlighting key capabilities but not whether models can reason about and enforce visualization principles within chart specifications.
For instance, a model might answer a comprehension question correctly yet still generate a chart specification that encodes ordered data with color hue (see Figure ~\ref{fig:figure_1})—an explicit principle violation~\cite{cleveland1984graphical} .

In this paper, we ask: \emph{Do large language models understand data visualization principles?}
We investigate whether LLMs—both text-only and multimodal—can interpret chart specifications, reason about established principles from perception and design research, and accurately detect violations of these principles.
To this end, we translate a subset of \emph{Draco}’s ASP constraints~\cite{moritz2018formalizing} into natural-language principles, generate a controlled dataset of 2,000 Vega-Lite~\cite{satyanarayan2016vega} specifications annotated with explicit violations, and evaluate well-known models on detecting noncompliance with data visualization principles.
To our knowledge, this is the first study to evaluate LLMs against formally specified visualization principles encoded as ASP constraints, and to bridge the gap between rule-based solvers and image-only evaluations. 

To guide our analysis, in this paper we focus on three research questions:

\begin{itemize}
\item \textbf{Q1:} \textit{Can LLM models assess whether charts adhere to visualization principles?}
\item \textbf{Q2:} \textit{Do multimodal VLMs (image + text) outperform text-only models when reasoning about design principles?}
\item \textbf{Q3:} \textit{Can LLMs enforce visualization principles to improve chart quality?}
\end{itemize}

To address these questions, we evaluate model performance on both the \emph{checking} and \emph{fixing} tasks. 
Gemini-2.5-Flash achieves the strongest checking results (F1 = 0.678 in the text-only setting and 0.716 with multimodal input) and the highest fixing performance (94.3\% enforcement rate). 
However, models still struggle with subtler perceptual constraints, where F1 scores drop below \textless\,0.10.
These results underscore both the promise and current limits of LLMs as language-based evaluators of visualization principles.
By evaluating detection and correction abilities, our study benchmarks the key functions required by visualization linters, laying the groundwork for tools that assist authors in identifying and improving data visualization issues.

\section{Related Work}
Understanding how large (vision-)language models reason about visualization principles requires connecting several research threads—from low-level chart data extraction to higher-level reasoning and design evaluation.
We review four complementary areas of research: (1) visualization data extraction, (2) reasoning with vision-language models (VLMs), (3) visualization linting, and (4) interactive assistance in visualization authoring.

\paragraph{Visualization Data Extraction.}
In order to detect and analyze visualization problems, early systems first needed to recover the information encoded in charts, both the underlying data and the visual encodings used to display them.
Before the advent of VLM-based approaches, this task relied on classical computer vision techniques.
Early efforts such as ReVision~\cite{savva2011revision} and ChartSense~\cite{jung2017chartsense} employed rule-based heuristics and classical machine learning to segment chart components and reconstruct underlying data from bitmap images.
Later approaches advanced this goal using neural architectures and end-to-end systems that convert chart images into structured outputs like data tables or programmatic specifications~\cite{ganguly2021systematic,luo2021chartocr, liu2022deplot, xue2023chartdetr, chen2024chart2vec,zhao2025chartcoder}.
However, recovering data and design decisions alone does not constitute understanding: reasoning about how visual encodings convey information or comply with design principles requires models that move beyond perceptual reconstruction toward semantic and normative interpretation.

\paragraph{Visualization Reasoning with LLMs}
The rapid emergence of large vision-language models (VLMs) has sparked interest in whether these systems can reason about visualization quality and design integrity.
The first attempts to probe this capability focused on visual reasoning over plots and charts, introducing datasets such as PlotQA~\cite{methani2020plotqa} and ChartQA~\cite{masry2022chartqa}, which benchmark question answering and logical inference over structured graphical data.
Lo et al.~\cite{lo2024good} demonstrated that multimodal LLMs can detect a broad range of misleading chart patterns directly from images, covering 21 issue categories.
Building on these efforts, Alexander et al.~\cite{alexander2024can} evaluated GPT-4 variants on real-world visualization pairs, showing that performance improves when prompts include explicit definitions and illustrative examples.
Other authors extend this line of inquiry by examining the broader notion of \emph{visualization literacy}: the ability of models to read, interpret, and reason about visual representations.
Hong et al.~\cite{hong2025llms} measures how well language models interpret charts, revealing that they often rely on prior knowledge rather than the visual information itself.
Pandey and Ottley~\cite{pandey2025benchmarking} conducted a standardized evaluation using VLAT and CALVI~\cite{ge2023calvi}, showing that models perform reliably on basic chart interpretation but struggle with misleading or complex visualizations.
Valentim et al.~\cite{valentim2025plot} extended this work by systematically varying chart type, title, and color, demonstrating that design features strongly affect model comprehension and accuracy.
Other recent benchmarks~\cite{iyengar2025interchartbenchmarkingvisualreasoning,tang2025chartmuseumtestingvisualreasoning,mukherjee2025encqabenchmarkingvisionlanguagemodels} confirm that current LLMs still lack robust visual reasoning, often overestimating their interpretive competence.

Together, these studies show that while models can spot obvious chart errors or answer basic comprehension questions, their understanding remains shallow and dependent on prior knowledge rather than visual evidence.
In contrast, our work targets specification-level principle verification—testing whether LLMs can detect explicit design-rule violations grounded in formal constraints.

\paragraph{Visualization Linters.}
Parallel to model-based reasoning, visualization researchers have long sought to formalize design knowledge into constraint-based systems capable of automatically checking charts for rule violations.
Moritz et al.~\cite{moritz2018formalizing} introduced \emph{Draco}, which encodes design guidelines as logical constraints and verifies chart specifications against them.
Building on this line of work, Chen et al.~\cite{chen2021vizlinter}, Hopkins et al.~\cite{hopkins2020visualint}, and McNutt et al.~\cite{mcnutt2018linting} introduce visualization \emph{linters}: linter–fixer pipelines that analyze either chart images or their underlying specifications to automatically flag and, in some cases, repair visualization problems. 
All of these linter frameworks ultimately rely on rule sets that are manually defined by experts. Each constraint encodes expert knowledge about visualization best practices—when to use stacking, color encoding, or axis scaling—and requires explicit implementation and ongoing maintenance as design insights evolve. 
While this symbolic approach ensures interpretability and precise control, it also limits scalability: expanding coverage to new visualization types or perceptual findings requires substantial manual effort.
Thus, while highly accurate as verification engines, these systems lack the flexibility and generalization that data-driven approaches can offer.

\paragraph{LLMs as Visualization Co-Designers.}
Beyond static critique, recent work explores LLMs as interactive partners in the visualization authoring process.
Shin et al.~\cite{shin2025visualizationary} embedded model feedback directly into design workflows, providing guideline-based prompts that suggest actionable improvements in real time.
Shen et al.~\cite{shen2024ask} examined troubleshooting scenarios, comparing human- and AI-assisted support during visualization development.
They found that humans provide nuanced, context-aware feedback, whereas LLMs respond faster but are often less grounded in the specific context.
VisJudge-Bench~\cite{xie2025visjudgebenchaestheticsqualityassessment} introduced an expert-annotated, image-only benchmark for assessing visualization fidelity, expressiveness, and aesthetics, and proposed a domain-tuned model trained on this dataset to better align with human quality judgments.
Together, these studies highlight the potential of LLMs and VLMs as collaborative agents in visualization design, capable of assisting users through iterative feedback and evaluation.
In contrast, our work does not aim to enable co-design but to formally assess whether LLMs possess the knowledge required for it, specifically, their ability to recognize, reason about, and correct visualizations in a structured setting.
\section{Benchmark Datasets}
\begin{figure*}[th]
    \centering
    \includegraphics[width=1\linewidth]{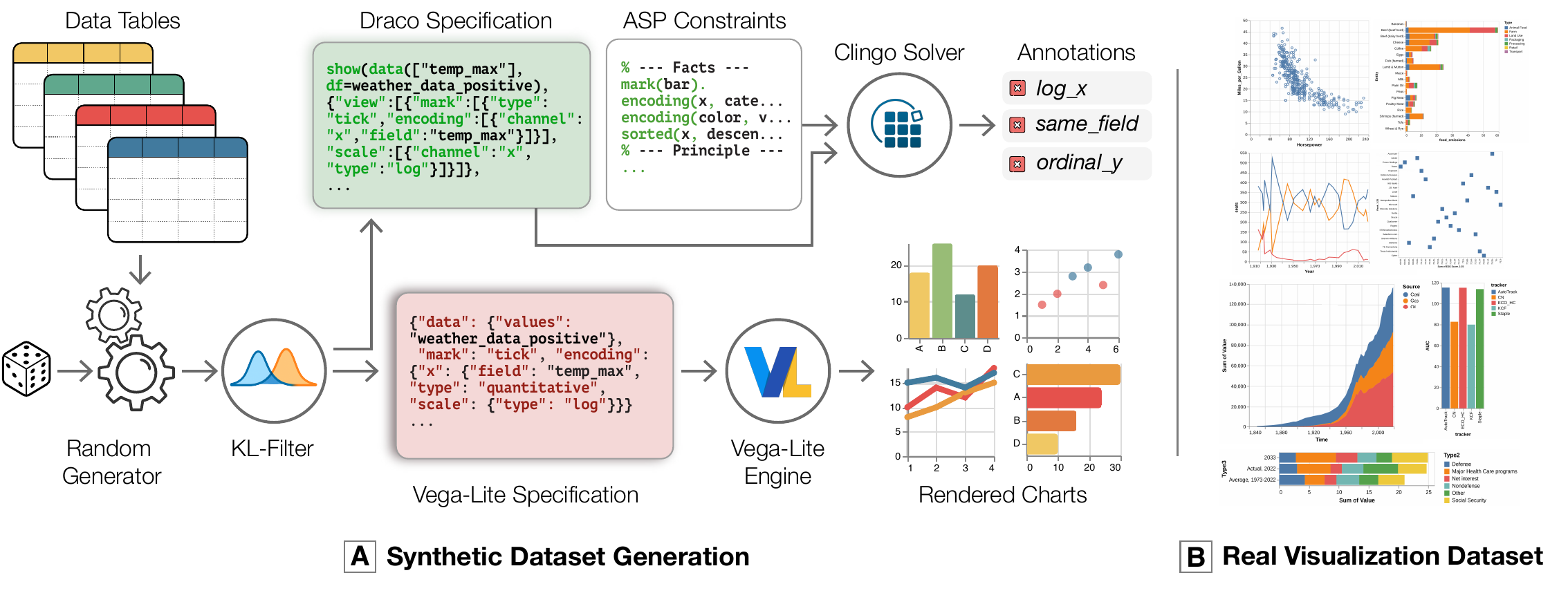}
    
    \caption{
\textbf{Overview of the benchmark datasets and generation pipeline.}
\textbf{(A)} The \emph{synthetic dataset} is produced by sampling \emph{Draco} chart specifications from tabular data, automatically generating Vega-Lite renderings and ground-truth annotations of design principle violations.
A Kullback–Leibler (KL) divergence filter ensures balanced coverage across violation types by accepting only candidates that improve uniformity in the distribution of problems.
\textbf{(B)} The \emph{real visualization dataset} complements the synthetic corpus by translating human-authored Vega-Lite specifications from GitHub into \emph{Draco} grammar, enabling principle-level analysis of authentic visualization practices.}
    
    \label{fig:figure_2}
\end{figure*}
While some datasets of misleading charts already exist~\cite{lo2022misinformed}, compiling over a thousand real-world examples and a taxonomy of 74 visual issues, they are not linked to formal design constraints or programmatic chart specifications.
Moreover, these collections consist mainly of “in-the-wild” images—photographs, screenshots, and editorial graphics, where chart elements are embedded in complex visual contexts. Such data are valuable for studying misinformation and visual framing but unsuitable for evaluating reasoning over design principles, as they conflate recognition and interpretation.
Therefore, we constructed two complementary datasets: one automatically generated to systematically cover a broad range of principle violations, and one manually curated from real examples that capture visualization practices and errors (see Figure~\ref{fig:figure_2}).

\subsection{Synthetic Dataset}
In order to generate synthetic charts, we began by selecting data tables from  thematic categories, including finance, health, and demographics, sourced from Kaggle~\footnote{\url{https://www.kaggle.com/}} (see Figure~\ref{fig:figure_2}, (\textbf{A})). 
We sanitize the dataset columns by removing spaces and non-alphanumeric characters using Python’s \emph{re} library~\footnote{\url{https://docs.python.org/3/library/re.html}}.
For each dataset, we employed a base chart specification expressed in the \emph{Draco} grammar, randomly assigning parameters such as \emph{mark type, encodings,} and \emph{variables} to produce varied yet valid visualization structures.
Rather than exhaustively enumerating all parameter combinations, we randomly sampled the chart configuration space.  

Random sampling alone tends to overrepresent a few common violation types, leading to skewed distributions.
To promote balance, we introduced a Kullback–Leibler (KL) divergence filter that retained only those candidate specifications improving coverage toward a uniform distribution of problems.
Let $p$ denote the empirical problem distribution and $u$ the uniform one; for each new candidate, we updated counts to obtain $p'$ and its divergence $\mathrm{KL}_{\text{new}} = \sum_i p_i \log \tfrac{p_i}{u_i}$.
Candidates were accepted if they reduced divergence by at least $\varepsilon = 10^{-3}$ or probabilistically according to Equation~\eqref{eq:acceptance}, with temperature $T=10^{-4}$:
\begin{equation}
\label{eq:acceptance}
P_{\text{accept}} = \exp\!\left(\frac{\Delta \mathrm{KL}}{T}\right),
\quad
\Delta \mathrm{KL} = \mathrm{KL}_{\text{old}} - \mathrm{KL}_{\text{new}}.
\end{equation}
This iterative process ensured steady progress toward balanced coverage while preventing convergence to local optima.

Each generated chart specification was converted into Vega-Lite format using a custom translation function and rendered.
Additionally, using \emph{Clingo}~\cite{kaminski2017clingo}, each specification was annotated by checking compliance with a subset of 57 ASP constraints derived from the original 206 described in \emph{Draco}, limited to cases that can be identified from the chart specification alone.
The final synthetic dataset comprises 2,000 Vega-Lite specifications derived from 21 source tables, annotated with 57 distinct principle (Appendix~\ref{app:problems}) types and a total of 12,858 instances of principle noncompliance.
Each principle category (e.g., color encoding of ordered data, redundant mappings, missing axes) is represented by roughly 200 samples.
The dataset also exhibits a diverse distribution of mark types, including \textit{rect} (906), \textit{line} (846), \textit{point} (1164), \textit{tick} (780), \textit{bar} (1011), \textit{area} (1185), and \textit{arc} (87) charts, with some charts containing multiple mark types.

\subsection{Real Visualization Dataset}
In addition to our synthetic corpus, we incorporate the Vega-Lite dataset introduced by Ko et al.~\cite{kovega}.
This collection originally contains 1,981 human-authored Vega-Lite specifications gathered from public GitHub repositories  (see Figure~\ref{fig:figure_2}, (\textbf{B})).
To enable visualization principle analysis, we automatically translated these Vega-Lite specifications into \emph{Draco} grammar using a custom converter function, preserving all encoding channels, marks, and data mappings.
During this process, 307 of the specifications were successfully converted, while 1,674 were discarded due to syntax incompatibilities, missing data references, or unsupported features such as multi-view layouts.
Each translated specification was first sanitized by removing spaces and non-alphanumeric characters from all column names using Python’s regular-expression library, and then automatically checked for principle violations using the same ASP constraint set applied to the synthetic corpus.
Although all 57 rules were tested, only 16 were triggered across the real dataset.
Consequently, the evaluation metrics reported on the real dataset partition in the following sections are computed over this reduced subset of principles.

Since this dataset is smaller than the synthetic corpus, we did not apply any balancing or filtering procedure to equalize the frequency of principle violations.
As a result, some violation types appear much more frequently than others, reflecting the natural distribution of design issues found in practice.
Unlike template-based synthetic datasets, these specifications exhibit more diversity and complexity.
The dataset thus captures authentic design practices found in the wild, providing a valuable complement to our controlled synthetic dataset.

\section{Q1: Can LLM Models Assess Whether Charts Adhere to Visualization Principles?}
\label{sec:llms}

\begin{figure}[t]
    \centering
    \includegraphics[width=1\linewidth]{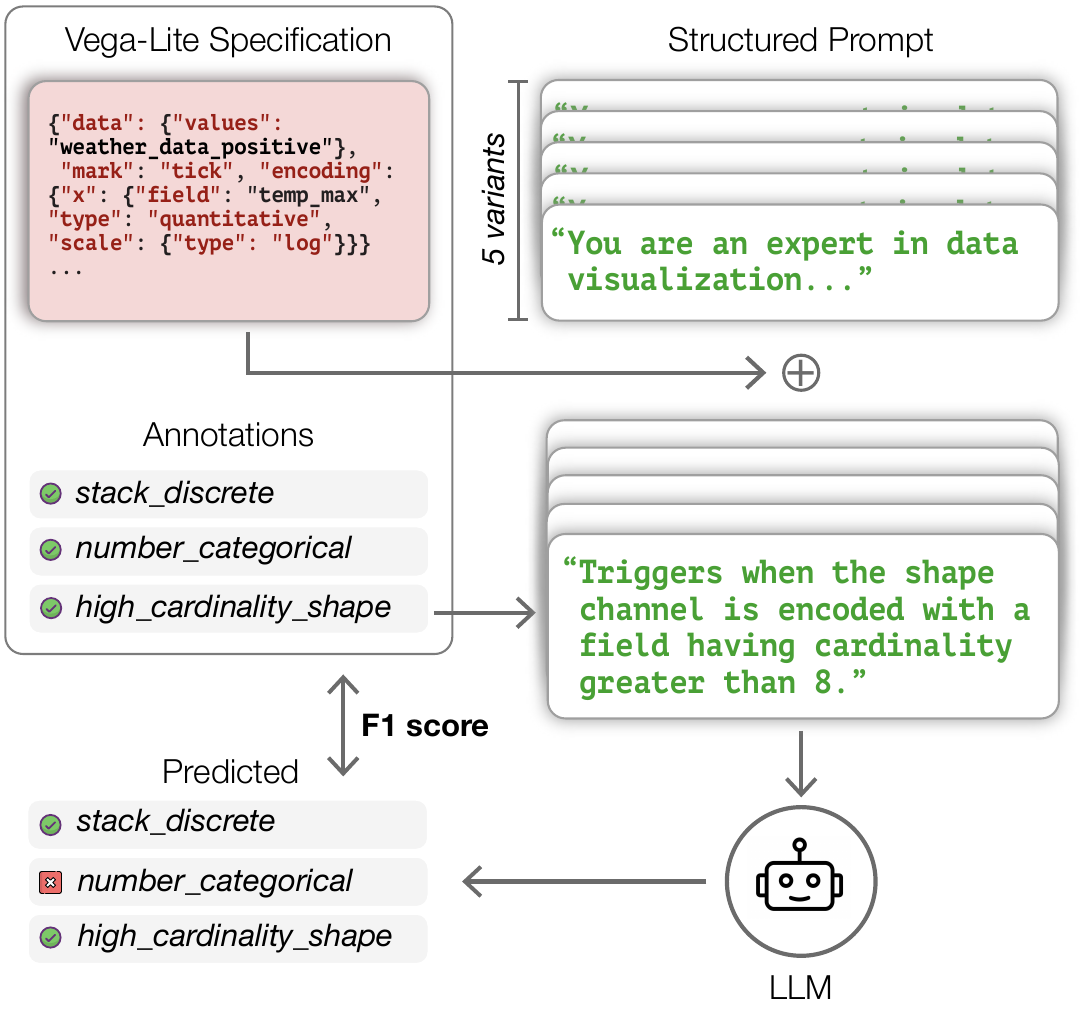}
    \caption{
    \textbf{Evaluation setup for assessing LLM understanding of visualization principles.}
    Each chart instance is defined by its Vega-Lite specification and annotated with one or more principle violations (e.g., \textit{high\_cardinality\_shape}, \textit{stack\_discrete}, \textit{number\_categorical}).
    Structured prompts (five semantically equivalent variants per principle) are used to query each model.
    Model outputs, expressed in a fixed JSON schema, are validated and compared to the reference annotations to compute F1-scores across all principles.}

    \label{fig:figure_3}
\end{figure}

Using our annotated datasets, we first evaluated the ability of large language models to identify visualization design issues directly from textual Vega-Lite specifications (see Figure~\ref{fig:figure_3}).

\subsection{Evaluation Setup}
To ensure consistency and reproducibility, we designed a structured prompting protocol. 
Each prompt included:
(i) a role description (e.g., \textit{You are an expert in data visualization design ...}); 
(ii) a list of target visualization principles to check, categorized in 5 groups; \emph{encoding}, \emph{marks}, \emph{stack}, \emph{scale}, \emph{data} (see Appendix \ref{app:VizGroups} for description); and 
(iii) the complete Vega-Lite specification, detailing mark type, encodings, channels, variables, and aggregations, along with up to 50 rows of its underlying data table.
Model responses were required to follow a fixed JSON format specifying which principles were triggered, enabling automatic parsing and metric computation.
Because each chart could involve multiple principles, we evaluated models using macro-averaged F1-scores computed over all principle categories. 
Results were averaged over three inference repetitions and five prompt variants per chart. 
The inclusion of standard deviation across repetitions captures both prompt sensitivity and model variability.

\subsection{Prompt Design}
To mitigate sensitivity to phrasing~\cite{alexander2024can}, we created five semantically equivalent prompt variants differing only in wording (see Appendix~\ref{app:prompts}).
For each inference, one variant was selected at random. 
All outputs were then checked using Pydantic~\footnote{\url{https://docs.pydantic.dev/latest/}} to validate the JSON structure.
Because ASP constraints express principles in a symbolic form unfamiliar to LLMs, each \emph{Draco} constraint was paraphrased into clear natural-language statements (see Appendix~\ref{app:NLprinciples}). 
We used ChatGPT 5.0 to generate readable reformulations, which were reviewed and refined by a member of the team to preserve semantic equivalence with the original rules.

\subsection{Computational Setup}
We accessed open-source models through Ollama~\footnote{\url{https://ollama.com/}}.
All experiments were conducted either on a Vast.ai~\footnote{\url{https://vast.ai/}} instance equipped with an AMD Ryzen Threadripper 2950X 16-Core Processor, two NVIDIA RTX 3090 GPUs, and 64 GB of RAM, or on our in-house server featuring an Intel Xeon Silver 4310 CPU @ 2.10GHz, 256 GB of RAM, and two NVIDIA A100 80GB GPUs. 
These configurations provided sufficient computational resources to support multiple inference rounds efficiently. 
Gemini-2.5-flash and GPT-4o were run using their respectives APIs. The complete run time of all models was approximately 480 hours.

\subsection{Results}

We evaluated each model’s ability to detect deviations from visualization principles across both the synthetic and real benchmark datasets.
F1-scores were computed per principle and macro-averaged to capture overall model performance, with standard deviations reflecting sensitivity to prompt variation and stochasticity across runs.
Table~\ref{tab:table_1} summarizes the results for all evaluated models.

On the synthetic dataset, scores were consistently modest across all models.
Among open-source models, GPT-OSS (20B) achieved the highest overall mean F1-score (0.580), followed by {Gemma-3-12B (0.324).
LLaMA, Llava, and Deepseek remained below 0.20.
Closed-source models performed better in this setting: Gemini-2.5-flash reached the highest score overall (0.678), with GPT-4o following at 0.529.
These results indicate that while large proprietary models outperform open counterparts, the F1 values remain moderate, suggesting that detecting visualization issues from text specifications remains a difficult task.

\sisetup{detect-weight=true,detect-inline-weight=math} 
\begin{table}[t]
\centering
\begin{threeparttable}
\setlength{\tabcolsep}{4pt}
\label{tab:open-closed-vlm-llm}
\begin{tabularx}{\columnwidth}{@{}l
S[table-format=1.3] S[table-format=1.3]
S[table-format=1.3] S[table-format=1.3]@{}}
\toprule
& \multicolumn{2}{c}{\textbf{Synthetic Dataset}} 
& \multicolumn{2}{c}{\textbf{Real Dataset}} \\
\cmidrule(lr){2-3} \cmidrule(lr){4-5}
\textbf{Model} & {\textbf{F1}↑} & {\textbf{STD}↓} 
               & {\textbf{F1}↑} & {\textbf{STD}↓} \\
\midrule
\addlinespace[2pt]
\textbf{Open-source} \\
Deepseek-v2-16B   &  0.172 & 0.005 & 0.202 & 0.009 \\
Llava-13B         &  0.174 & 0.005 & 0.182 & 0.007 \\
Llama-3.1-8B      &  0.195 & 0.009 & 0.217 & 0.015 \\
Gemma-3-12B       &  0.324 & 0.007 & 0.448 & 0.114 \\
GPT-OSS-20B       &  \bfseries 0.580 & 0.080 & \bfseries 0.701 & 0.060 \\
\midrule
\addlinespace[2pt]
\textbf{Closed-source} \\
GPT-4o            & 0.529 & 0.014 & 0.654 & 0.044 \\
Gemini-2.5-Flash  & \bfseries 0.678 & 0.016 & \bfseries 0.743 & 0.001 \\
\bottomrule
\end{tabularx}
\caption{\textbf{Comparison of open- and closed-source models across the \textit{Synthetic} and \textit{Real} datasets in text-only modality.}
We report mean F1 (↑) and standard deviation STD (↓) across runs. 
Best within each block in \textbf{bold}.}
\label{tab:table_1}
\end{threeparttable}
\end{table}

Performance was higher across all models on the real dataset.
Open-source models improved substantially: GPT-OSS rose to 0.701 and Gemma-3-12B to 0.448, while the best closed-source models, Gemini-2.5-flash and GPT-4o, reached 0.743 and 0.654, respectively.
In the real dataset, only a subset of 16 out of the 57 principles is evaluated, which naturally narrows the difficulty of the task.
In addition, because the charts come from public GitHub repositories, some of them may resemble patterns the models have encountered during pretraining, potentially making the problem more familiar than in the synthetic setting.
This highlights the importance of complementing real-world benchmarks with synthetic examples that better isolate genuine reasoning and generalization abilities.


Figure~\ref{fig:figure_4} (\textbf{A}) shows macro F1-scores by mark type in the synthetic dataset.
Across all models, \textit{arc} marks are consistently the most difficult, yielding the lowest scores.
Bar, line, and point marks are generally easier, with higher and more stable performance across models.
These differences highlight that certain visual encodings pose systematically greater challenges for current LLMs.
Figure~\ref{fig:figure_4} (\textbf{B}) displays macro F1-scores aggregated by principle category and reveals a clear two-group structure.
The top-performing models (Gemini-2.5-Flash, GPT-OSS-20B, and GPT-4o) exhibit highly similar trends across categories, maintaining strong and consistent accuracy.
In contrast, the four lower-performing models follow a parallel but uniformly weaker pattern.
This separation suggests that models cluster into two regimes: within each group they share similar patterns of strengths and weaknesses across principles, while these patterns differ systematically between groups.

\begin{figure}[t]
    \centering
    \includegraphics[width=1\linewidth]{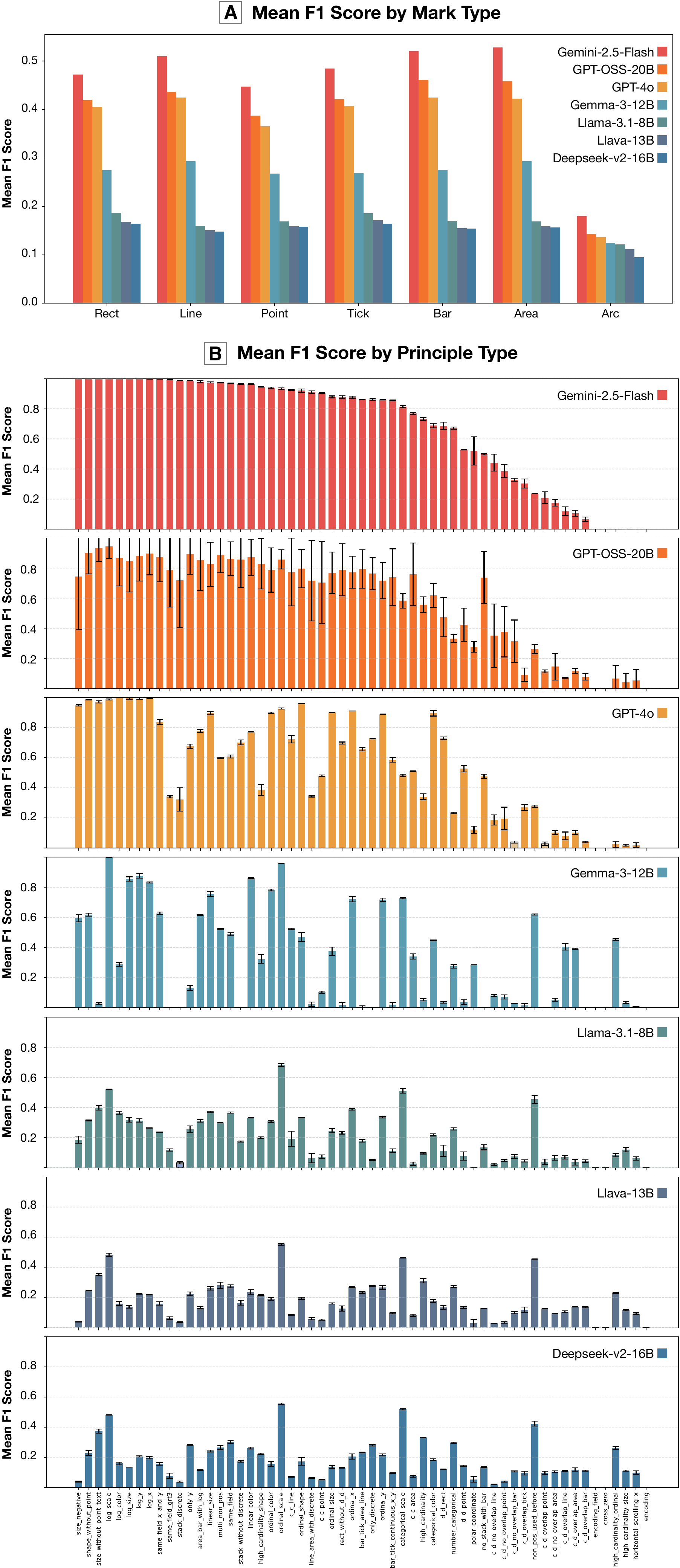}
    \caption{
\textbf{Model performance across mark types and problem categories.}
\textbf{(A)} Mean F1-scores by \emph{mark type}, showing how different visual encodings influence model accuracy.
\textbf{(B)} Mean F1-scores by \emph{principle} sorted by Gemini-2.5-Flash performance.
}
    \label{fig:figure_4}
\end{figure}

Overall, these results reveal a marked difference in how models handle familiar versus systematically generated visualization structures.
While state-of-the-art LLMs can recognize and classify common design violations, their reasoning over abstract or atypical principle configurations remains limited.

\section{Q2: Do Multimodal VLMs (Image + Text) Outperform Text-Only Models When Reasoning About Design Principles?}

In the previous section, we evaluated how well LLMs can detect violations of visualization principles when provided only with the Vega-Lite specification and a subset of its underlying data.
We now extend this analysis to multimodal vision–language models and examine whether access to the rendered chart image improves their ability to evaluate adherence to visualization principles.
In particular, we ask whether incorporating the visual representation enhances a model’s capacity to identify issues beyond what can be inferred from the declarative chart specification alone.

\subsection{Evaluation Setup}
The evaluation protocol mirrors the text-only setup described in Section~\ref{sec:llms}, with the sole modification that each prompt additionally included the PNG rendering of the chart.
Images were generated using the \emph{vl-convert}~\footnote{\url{https://pypi.org/project/vl-convert-python/}} rendering engine at a resolution of 800×600 pixels to ensure consistency across models.
Only models capable of processing image input were evaluated in this setting—namely Llava-13B, Gemma-3-12B, GPT-4o, and Gemini-2.5-Flash.
The remaining models from the text-only experiments were excluded as they do not support multimodal input.

Aside from the inclusion of the rendered image and a text line in the prompt indicating that an image will be issued, no component of the prompting pipeline was altered. 
Each model received the same role description, natural-language list of principles, and complete Vega-Lite specification, and was required to return a structured JSON response indicating which principles were triggered.
By keeping the evaluation procedure fixed, we isolate the contribution of visual input and directly quantify how much additional information VLMs extract from the chart’s appearance relative to its declarative specification.

\subsection{Results} 

Table~\ref{tab:table_2} summarizes the performance of multimodal VLMs compared with their text-only counterparts.
On the synthetic dataset, multimodal input leads to modest but consistent improvements.
Gemma-3-12B remains the strongest open-source model (0.319), and both GPT-4o and Gemini-2.5-Flash outperform their text-only versions, with Gemini-2.5-Flash achieving the highest score overall (0.716).
These gains, although measurable, remain limited, suggesting that current models struggle to fully exploit the additional visual signal, even though all principle violations are visually apparent in the chart.

On the real dataset, improvements are slightly larger.
Gemma-3-12B, GPT-4o, and Gemini-2.5-Flash all benefit from the addition of the chart image, with Gemini-2.5-Flash again obtaining the best performance (0.778).
Taken together, these results indicate that multimodal input is beneficial, but they also point to substantial headroom for models that more effectively exploit the combined signals from text and image.

\begin{table}[t]
\centering
\begin{threeparttable}
\setlength{\tabcolsep}{4pt}
\label{tab:open-closed-vlm-llm}
\begin{tabularx}{\columnwidth}{@{}l
S[table-format=1.2] S[table-format=1.2]
S[table-format=1.2] S[table-format=1.2]@{}}
\toprule
& \multicolumn{2}{c}{\textbf{Synthetic Dataset}} 
& \multicolumn{2}{c}{\textbf{Real Dataset}} \\
\cmidrule(lr){2-3} \cmidrule(lr){4-5}
\textbf{Model} & {\textbf{F1}↑} & {\textbf{STD↓}} 
               & {\textbf{F1}↑} & {\textbf{STD↓}} \\
\midrule
\addlinespace[2pt]
\textbf{Open-source} \\
Llava-13B           & 0.172 & 0.015 & 0.184 & 0.008 \\
Gemma-3-12B         & \textbf{0.319} & 0.012 & \textbf{0.460} & 0.153 \\
\midrule
\addlinespace[1pt]
\textbf{Closed-source} \\
GPT-4o         & 0.564 & 0.018 & 0.657 & 0.020 \\
Gemini-2.5-Flash   & \textbf{0.716} & 0.041 & \textbf{0.778} & 0.030 \\
\bottomrule
\end{tabularx}
\caption{\textbf{Comparison of open- and closed-source models across the \textit{Synthetic} and \textit{Real} datasets in text+image modality.}
We report mean F1 (↑) and standard deviation STD (↓) across runs. 
Best within each block in \textbf{bold}.}
\label{tab:table_2}
\end{threeparttable}
\end{table}

\section{Q3: Can LLMs Enforce Visualization Principles to Improve Chart Quality?}

Previously, we focused on whether both LLMs and VLMs were capable of detecting visualization principles directly from chart specifications. 
However, detection alone does not fully capture a model’s usefulness in practical visualization workflows. 
Beyond simply identifying design issues, it becomes equally important (and arguably more challenging) to determine whether these models can modify a specification. 
In other words, can models not only recognize a principle non-compliance but also generate a corrected version of the chart that successfully satisfies it? 
Exploring this question allows us to evaluate the models capacity for targeted repair and specification editing, thereby extending our analysis from passive detection to active problem fixing.
\subsection{Evaluation Setup}

Our evaluation pipeline proceeds as follows.
For each chart in the dataset, we randomly select one of its associated violated principles.
The model is then prompted to generate a corrected Vega-Lite specification that satisfies the target principle while making the minimal necessary modifications to the original specification (see Appendix~\ref{app:fixing_prompt} for the full prompt).
The model’s output is subsequently translated back into the \emph{Draco} grammar and automatically checked to determine the updated set of principle violations.

To assess the ability of LLMs to correct visualization principles, we selected the best-performing open-source (GPT-OSS-20B) and closed-source (Gemini-2.5-Flash) models from Section~\ref{sec:llms}.
We evaluated 1,827 synthetic specifications, each paired with a randomly sampled violated principle.
We then computed several complementary metrics to measure not only whether the targeted violation is successfully resolved but also whether the model inadvertently introduces new infractions.

\paragraph{Compilability (CO).}
The percentage of generated Vega-Lite specifications that can be successfully rendered, indicating whether the model produces syntactically valid code.

\paragraph{Enforcement Rate (ER).}
The success rate with which the model resolves the targeted violation.
For each generated specification, we verify whether the selected principle is absent from the updated list of violations.

\paragraph{Compliance Ratio (CR).}
A measure of how the overall number of violations changes after correction.
It is defined as $CR = P_{\text{fixed}} / P_{\text{original}}$, 
where $P_{\text{original}}$ is the number of violations in the original specification 
and $P_{\text{fixed}}$ the number in the corrected one.
Values below 1 indicate a net reduction in violations, whereas values above 1 reflect an increase.
This metric captures the model’s broader impact on the specification’s compliance, beyond the targeted principle.


\subsection{Results}
Table~\ref{tab:fixing-llm} reports the compilability, enforcement rate, and compliance ratio for the two evaluated models.
Both LLMs demonstrate a strong ability to generate syntactically valid Vega-Lite specifications: GPT-OSS-20B achieves a compilability of 0.99, while Gemini-2.5-Flash produces valid output in the vast majority of cases as well.
This indicates that, at least for the synthetic specifications considered, both models are capable of producing structurally coherent chart definitions when asked to apply targeted corrections.

In terms of correcting the selected violation, Gemini-2.5-Flash achieves the highest Enforcement Rate (94.3\%), outperforming GPT-OSS-20B (86.3\%).
The compliance ratio is similar for both models (0.724 for GPT-OSS-20B and 0.724 for Gemini-2.5-Flash), indicating that their broader impact on overall specification quality is nearly equivalent: both reduce, on average, roughly 28\% of the total violations present in the original charts.
Taken together, these results show that large models can perform targeted repairs with relatively high success while also improving overall compliance.
However, neither model consistently makes selective edits that enforce a single principle without affecting others, indicating substantial room for improvement in precise, constraint-aware specification editing.

\begin{table}[t]
\centering
\begin{tabularx}{\columnwidth}{lXXX}
\toprule
 & \multicolumn{3}{c}{\textbf{Synthetic Dataset}} \\
\cmidrule(lr){2-4}
\textbf{Model} & \textbf{CO}\,$\uparrow$ & \textbf{ER}\,$\uparrow$ & \textbf{CR} \\
\midrule
\textbf{Open-source} & & & \\
GPT-OSS-20B & 99\% & 86.3\% & \textbf{0.722} \\
\midrule
\textbf{Closed-source} & & & \\
Gemini-2.5-Flash & 100\% & \textbf{94.3\%} & 0.724 \\
\bottomrule
\end{tabularx}

\caption{
\textbf{Comparison of open- and closed-source models across
the Synthetic dataset in text-only modality}. We report
mean Enforcement Rate (ER↑) and Compliance Ratio (CR) across runs.}
\label{tab:fixing-llm}
\end{table}


\section{Conclusion}
In this work, we presented the first systematic evaluation of LLMs and VLMs for detecting visualization design principles directly from Vega-Lite specifications and images.
We introduced a dataset of 2{,}000 instances covering 57 principles and a broad range of non-compliances, together with standardized prompts and evaluation metrics, and evaluated both open- and closed-source models on synthetic and real benchmarks.

Our results show that large proprietary models currently lead performance, but the task remains challenging. 
On the synthetic dataset, even the strongest text-only model reaches only moderate accuracy (F1 = 0.678), and VLMs improve this lightly (up to F1 = 0.716). 
Performance increases on the real dataset (up to F1 = 0.778), but this gain reflects a narrower set of principles and likely exposure to similar patterns during pretraining. 
Fixing experiments reveal a notable asymmetry: models tend to be more effective at correcting violations than detecting them, with Gemini-2.5-Flash achieving a high 94.3\% enforcement rate despite modest detection performance. Together, these trends suggest that although current systems show promise, robust and principle-aware chart reasoning remains far from solved.

Looking ahead, several directions merit further study: (i) extending the benchmark to more chart types, tasks, and principle violations; (ii) systematically exploring VLM configuration and interaction settings (e.g., prompting strategies, multimodal input formats, and decoding parameters) to better exploit the joint text–image signal; and (iii) developing evaluation protocols that go beyond macro F1-score to assess partial correctness, calibration, and reasoning quality. Connecting these advances to downstream tools for chart auditing or visualization recommendation offers a promising path to practical impact.

\section{Acknowledgments} This project was supported by Universidad Torcuato Di Tella, Argentina and
Alfred P. Sloan Foundation, Grant G-2024-22665.

{
    \small
    \bibliographystyle{ieeenat_fullname}
    \bibliography{references}

@book{cleveland1984graphical,
  title={The elements of graphing data},
  author={Cleveland, William S},
  year={1985},
  publisher={Wadsworth Publ. Co.}
}

@article{hong2025llms,
  title={Do llms have visualization literacy? an evaluation on modified visualizations to test generalization in data interpretation},
  author={Hong, Jiayi and Seto, Christian and Fan, Arlen and Maciejewski, Ross},
  journal={IEEE Transactions on Visualization and Computer Graphics},
  year={2025},
  publisher={IEEE}
}

@book{huff1954lie,
  title={How to lie with statistics},
  author={Huff, Darrell},
  year={2023},
  publisher={Penguin UK}
}

@inproceedings{munzner2014visualization,
  title={Visualization analysis and design},
  author={Munzner, Tamara},
  booktitle={Proceedings of the Special Interest Group on Computer Graphics and Interactive Techniques Conference Courses},
  pages={1--2},
  year={2025}
}

@article{lo2024good,
  title={How good (or bad) are llms at detecting misleading visualizations?},
  author={Lo, Leo Yu-Ho and Qu, Huamin},
  journal={IEEE Transactions on Visualization and Computer Graphics},
  year={2024},
  publisher={IEEE}
}

@article{chen2021vizlinter,
  title        = {VizLinter: A Linter and Fixer Framework for Data Visualization},
  author       = {Chen, Qing and Sun, Fuling and Xu, Xinyue and Chen, Zui and Wang, Jiazhe and Cao, Nan},
  journal      = {IEEE Transactions on Visualization and Computer Graphics},
  volume       = {28},
  number       = {1},
  pages        = {206--216},
  year         = {2021}
}

@misc{kaminski2017clingo,
  title        = {Clingo},
  author       = {Kaminski, Roland},
  howpublished = {\url{https://github.com/potassco/clingo}},
  year         = {2017}
}

@article{moritz2018formalizing,
  title={Formalizing visualization design knowledge as constraints: Actionable and extensible models in draco},
  author={Moritz, Dominik and Wang, Chenglong and Nelson, Greg L and Lin, Halden and Smith, Adam M and Howe, Bill and Heer, Jeffrey},
  journal={IEEE transactions on visualization and computer graphics},
  volume={25},
  number={1},
  pages={438--448},
  year={2018},
  publisher={IEEE}
}

@misc{clingo,
  author       = {Roland Kaminski},
  title        = {Clingo},
  year         = {2017},
  howpublished = {\url{https://github.com/potassco/clingo}}
}

@book{healy2024data,
  title={Data visualization: a practical introduction},
  author={Healy, Kieran},
  year={2024},
  publisher={Princeton University Press}
}

@book{tufte1983visual,
  title={The visual display of quantitative information},
  author={Tufte, Edward R and Graves-Morris, Peter R},
  volume={2},
  number={9},
  year={1983},
  publisher={Graphics press Cheshire, CT}
}

@inproceedings{alexander2024can,
  title={Can gpt-4 models detect misleading visualizations?},
  author={Alexander, Jason and Nanda, Priyal and Yang, Kai-Cheng and Sarvghad, Ali},
  booktitle={2024 IEEE Visualization and Visual Analytics (VIS)},
  pages={106--110},
  year={2024},
  organization={IEEE}
}

@article{shin2025visualizationary,
  title={Visualizationary: Automating design feedback for visualization designers using llms},
  author={Shin, Sungbok and Hong, Sanghyun and Elmqvist, Niklas},
  journal={IEEE Transactions on Visualization and Computer Graphics},
  year={2025},
  publisher={IEEE}
}

@article{shen2024ask,
  title={Ask humans or ai? exploring their roles in visualization troubleshooting},
  author={Shen, Shuyu and Lu, Sirong and Shen, Leixian and Sheng, Zhonghua and Tang, Nan and Luo, Yuyu},
  journal={arXiv preprint arXiv:2412.07673},
  year={2024}
}

@misc{xie2025visjudgebenchaestheticsqualityassessment,
      title={VisJudge-Bench: Aesthetics and Quality Assessment of Visualizations}, 
      author={Yupeng Xie and Zhiyang Zhang and Yifan Wu and Sirong Lu and Jiayi Zhang and Zhaoyang Yu and Jinlin Wang and Sirui Hong and Bang Liu and Chenglin Wu and Yuyu Luo},
      year={2025},
      eprint={2510.22373},
      archivePrefix={arXiv},
      primaryClass={cs.CL},
      url={https://arxiv.org/abs/2510.22373}, 
}

@article{zhao2025chartcoder,
  title={Chartcoder: Advancing multimodal large language model for chart-to-code generation},
  author={Zhao, Xuanle and Luo, Xianzhen and Shi, Qi and Chen, Chi and Wang, Shuo and Liu, Zhiyuan and Sun, Maosong},
  journal={arXiv preprint arXiv:2501.06598},
  year={2025}
}

@article{masry2022chartqa,
  title={Chartqa: A benchmark for question answering about charts with visual and logical reasoning},
  author={Masry, Ahmed and Long, Do Xuan and Tan, Jia Qing and Joty, Shafiq and Hoque, Enamul},
  journal={arXiv preprint arXiv:2203.10244},
  year={2022}
}

@inproceedings{methani2020plotqa,
  title={Plotqa: Reasoning over scientific plots},
  author={Methani, Nitesh and Ganguly, Pritha and Khapra, Mitesh M and Kumar, Pratyush},
  booktitle={Proceedings of the ieee/cvf winter conference on applications of computer vision},
  pages={1527--1536},
  year={2020}
}

@inproceedings{luo2021chartocr,
  title={Chartocr: Data extraction from charts images via a deep hybrid framework},
  author={Luo, Junyu and Li, Zekun and Wang, Jinpeng and Lin, Chin-Yew},
  booktitle={Proceedings of the IEEE/CVF winter conference on applications of computer vision},
  pages={1917--1925},
  year={2021}
}

@article{liu2022deplot,
  title={Deplot: One-shot visual language reasoning by plot-to-table translation},
  author={Liu, Fangyu and Eisenschlos, Julian Martin and Piccinno, Francesco and Krichene, Syrine and Pang, Chenxi and Lee, Kenton and Joshi, Mandar and Chen, Wenhu and Collier, Nigel and Altun, Yasemin},
  journal={arXiv preprint arXiv:2212.10505},
  year={2022}
}

@inproceedings{ge2023calvi,
  title={Calvi: Critical thinking assessment for literacy in visualizations},
  author={Ge, Lily W and Cui, Yuan and Kay, Matthew},
  booktitle={Proceedings of the 2023 CHI conference on human factors in computing systems},
  pages={1--18},
  year={2023}
}

@inproceedings{lo2022misinformed,
  title={Misinformed by visualization: What do we learn from misinformative visualizations?},
  author={Lo, Leo Yu-Ho and Gupta, Ayush and Shigyo, Kento and Wu, Aoyu and Bertini, Enrico and Qu, Huamin},
  booktitle={Computer Graphics Forum},
  volume={41},
  number={3},
  pages={515--525},
  year={2022},
  organization={Wiley Online Library}
}

@article{lee2016vlat,
  title={Vlat: Development of a visualization literacy assessment test},
  author={Lee, Sukwon and Kim, Sung-Hee and Kwon, Bum Chul},
  journal={IEEE transactions on visualization and computer graphics},
  volume={23},
  number={1},
  pages={551--560},
  year={2016},
  publisher={IEEE}
}

@article{satyanarayan2016vega,
  title={Vega-lite: A grammar of interactive graphics},
  author={Satyanarayan, Arvind and Moritz, Dominik and Wongsuphasawat, Kanit and Heer, Jeffrey},
  journal={IEEE transactions on visualization and computer graphics},
  volume={23},
  number={1},
  pages={341--350},
  year={2016},
  publisher={IEEE}
}

@inproceedings{pandey2025benchmarking,
  title={Benchmarking visual language models on standardized visualization literacy tests},
  author={Pandey, Saugat and Ottley, Alvitta},
  booktitle={Computer Graphics Forum},
  pages={e70137},
  year={2025},
  organization={Wiley Online Library}
}

@book{ware2019information,
  title={Information visualization: perception for design},
  author={Ware, Colin},
  year={2019},
  publisher={Morgan Kaufmann}
}

@book{cairo2019charts,
  title={How charts lie: Getting smarter about visual information},
  author={Cairo, Alberto},
  year={2019},
  publisher={WW Norton \& Company}
}

@book{tufte1997visual,
  title={Visual explanations},
  author={Tufte, Edward R and Robins, David},
  year={1997},
  publisher={Graphics Cheshire, CT}
}

@inproceedings{savva2011revision,
  title={Revision: Automated classification, analysis and redesign of chart images},
  author={Savva, Manolis and Kong, Nicholas and Chhajta, Arti and Fei-Fei, Li and Agrawala, Maneesh and Heer, Jeffrey},
  booktitle={Proceedings of the 24th annual ACM symposium on User interface software and technology},
  pages={393--402},
  year={2011}
}

@inproceedings{jung2017chartsense,
  title={Chartsense: Interactive data extraction from chart images},
  author={Jung, Daekyoung and Kim, Wonjae and Song, Hyunjoo and Hwang, Jeong-in and Lee, Bongshin and Kim, Bohyoung and Seo, Jinwook},
  booktitle={Proceedings of the 2017 chi conference on human factors in computing systems},
  pages={6706--6717},
  year={2017}
}

@article{xue2023chartdetr,
  title={Chartdetr: A multi-shape detection network for visual chart recognition},
  author={Xue, Wenyuan and Chen, Dapeng and Yu, Baosheng and Chen, Yifei and Zhou, Sai and Peng, Wei},
  journal={arXiv preprint arXiv:2308.07743},
  year={2023}
}

@article{chen2024chart2vec,
  title={Chart2vec: A universal embedding of context-aware visualizations},
  author={Chen, Qing and Chen, Ying and Zou, Ruishi and Shuai, Wei and Guo, Yi and Wang, Jiazhe and Cao, Nan},
  journal={IEEE Transactions on Visualization and Computer Graphics},
  volume={31},
  number={4},
  pages={2167--2181},
  year={2024},
  publisher={IEEE}
}

@inproceedings{ganguly2021systematic,
  title={A systematic evaluation of object detection networks for scientific plots},
  author={Ganguly, Pritha and Methani, Nitesh S and Khapra, Mitesh M and Kumar, Pratyush},
  booktitle={Proceedings of the AAAI Conference on Artificial Intelligence},
  volume={35},
  number={2},
  pages={1379--1387},
  year={2021}
}

@article{valentim2025plot,
  title={The Plot Thickens: Quantitative Part-by-Part Exploration of MLLM Visualization Literacy},
  author={Valentim, Matheus and Dhanoa, Vaishali and Le{\'o}n, Gabriela Molina and Elmqvist, Niklas},
  journal={arXiv preprint arXiv:2504.02217},
  year={2025}
}

@misc{mukherjee2025encqabenchmarkingvisionlanguagemodels,
      title={EncQA: Benchmarking Vision-Language Models on Visual Encodings for Charts}, 
      author={Kushin Mukherjee and Donghao Ren and Dominik Moritz and Yannick Assogba},
      year={2025},
      eprint={2508.04650},
      archivePrefix={arXiv},
      primaryClass={cs.CV},
      url={https://arxiv.org/abs/2508.04650}, 
}

@misc{iyengar2025interchartbenchmarkingvisualreasoning,
      title={InterChart: Benchmarking Visual Reasoning Across Decomposed and Distributed Chart Information}, 
      author={Anirudh Iyengar Kaniyar Narayana Iyengar and Srija Mukhopadhyay and Adnan Qidwai and Shubhankar Singh and Dan Roth and Vivek Gupta},
      year={2025},
      eprint={2508.07630},
      archivePrefix={arXiv},
      primaryClass={cs.CL},
      url={https://arxiv.org/abs/2508.07630}, 
}

@misc{tang2025chartmuseumtestingvisualreasoning,
      title={ChartMuseum: Testing Visual Reasoning Capabilities of Large Vision-Language Models}, 
      author={Liyan Tang and Grace Kim and Xinyu Zhao and Thom Lake and Wenxuan Ding and Fangcong Yin and Prasann Singhal and Manya Wadhwa and Zeyu Leo Liu and Zayne Sprague and Ramya Namuduri and Bodun Hu and Juan Diego Rodriguez and Puyuan Peng and Greg Durrett},
      year={2025},
      eprint={2505.13444},
      archivePrefix={arXiv},
      primaryClass={cs.CL},
      url={https://arxiv.org/abs/2505.13444}, 
}

@inproceedings{hopkins2020visualint,
  title={Visualint: Sketchy in situ annotations of chart construction errors},
  author={Hopkins, Aspen K and Correll, Michael and Satyanarayan, Arvind},
  booktitle={Computer Graphics Forum},
  volume={39},
  number={3},
  pages={219--228},
  year={2020},
  organization={Wiley Online Library}
}

@inproceedings{mcnutt2018linting,
  title={Linting for visualization: Towards a practical automated visualization guidance system},
  author={McNutt, Andrew and Kindlmann, Gordon},
  booktitle={VisGuides: 2nd Workshop on the Creation, Curation, Critique and Conditioning of Principles and Guidelines in Visualization},
  volume={1},
  number={2},
  pages={9},
  year={2018}
}

@article{kovega,
  title={A Vega-Lite Dataset and Natural Language Generation Pipeline with Large Language Models},
  author={Ko, Hyung-Kwon and Jeon, Hyeon and Park, Gwanmo and Kim, Dae Hyun and Kim, Nam Wook and Kim, Juho and Seo, Jinwook}
}
}

\clearpage

\section{Appendix}

\subsection{Prompts}
\label{app:prompts}

We provide the exact prompt templates used in our experiments. Each template contains placeholders that are replaced at runtime: \texttt{\{vega-spec\}} denotes the Vega-Lite specification under evaluation, and \texttt{\{problems\}} corresponds to the full list of candidate problem names considered in our study. 

\lstset{
  basicstyle=\ttfamily\small,
  breaklines=true,
  frame=single,
  columns=fullflexible
}

\textbf{Prompt \#1}
\begin{lstlisting}[language=]
You are an expert in data visualization design using Vega-Lite.

## Input

### Problems:
{problems}

### Vega-Lite specification:
{vega-spec}

## Task
Analyze the Vega-Lite specification and identify which **exact** problem name from the list above are present.

## Output Requirements
- Output **only** a valid JSON array of strings.
- Each string must be an **exact match** to a problem name from the provided list (excluding the "name :" prefix).
- Do **not** add explanations, reasoning, or any extra text.
- If no problems match, return an empty JSON array: []

### Example:
["problem_A", "problem_B"]
\end{lstlisting}

\textbf{Prompt \#2}
\begin{lstlisting}[language=]
### Problems:
{problems}

You are a Vega-Lite visualization evaluator. Your goal is to read the given Vega-Lite specification and identify any problems from the above list that it exhibits. Focus only on exact matches from the provided names.

### Vega-Lite specification:
{vega-spec}

## Output Requirements
- Output **only** a JSON array of strings.
- Strings must exactly match a problem name from the list (omit "name :").
- No explanations, commentary, or extra formatting.
- If no problems are present, return []

### Example:
["problem_A", "problem_B"]
"""
\end{lstlisting}

\textbf{Prompt \#3}
\begin{lstlisting}[language=]
### Problems:
{problems}

Analyze the following Vega-Lite specification carefully. Report which of the problem names listed above are present in it. Only use exact matches.

### Vega-Lite specification:
{vega-spec}

## Output Requirements
- Return **only** a JSON array of strings.
- Each string must match a problem name exactly (exclude "name :").
- Do not include explanations, notes, or any additional text.
- Return [] if there are no matches.

### Example:
["problem_A", "problem_B"]
"""
\end{lstlisting}

\textbf{Prompt \#4}
\begin{lstlisting}[language=]
### Problems:
{problems}

You are tasked with checking the Vega-Lite specification below. Identify all problems from the above list that appear in it. Ensure each match is exact.

### Vega-Lite specification:
{vega-spec}

## Output Requirements
- Provide **only** a JSON array of strings.
- Each string must be an exact problem name (without "name :").
- No extra text, reasoning, or commentary.
- If no problems are found, return []

### Example:
["problem_A", "problem_B"]
"""
\end{lstlisting}

\textbf{Prompt \#5  }
\begin{lstlisting}[language=]
### Problems:
{problems}

Review the Vega-Lite specification and determine which problem names from the list are present. Only include exact matches in your output.

### Vega-Lite specification:
{vega-spec}

## Output Requirements
- Output **only** a JSON array of strings.
- Strings must exactly match the problem names (ignore the "name :" prefix).
- Do not provide explanations, notes, or any additional text.
- If none match, return []

### Example:
["problem_A", "problem_B"]
"""
\end{lstlisting}

\label{app:fixing_prompt}
\subsection{Fixing Prompt}

\label{app:problems}

\subsection{Natural Language Principles}

The handcrafted ASP \emph{Draco} constraints were translated into natural language using ChatGPT 5.0 to produce readable reformulations, which were subsequently reviewed and refined by a member of the research team. Below we present the complete list of principles in the exact form provided to the LLMs and VLMs.

\begin{lstlisting}[language=]
{'size_negative': 'A violation occurs if a channel is encoded with size and the corresponding field contains both negative and positive values. The size encoding implies positive magnitude, so it should not be used when the data includes negative values.',
 'line_area_with_discrete': 'A violation occurs when a line or area chart is used and both the x and y channels are encoded with discrete scales. Line and area marks are intended for continuous data and do not function correctly with fully discrete axes.',
 'bar_tick_continuous_x_y': 'A violation occurs if a bar or tick chart is used and both the x and y channels are continuous. These mark types are designed to compare discrete categories and are not suitable for continuous to continuous configurations.',
 'shape_without_point': 'A violation occurs when the shape channel is used but the mark type is not point. The shape encoding is only meaningful when applied to point marks.',
 'size_without_point_text': 'A violation occurs if the size channel is used with a mark type that is neither point nor text. Only point and text marks properly support the size encoding.',
 'area_bar_with_log': 'A violation occurs when a bar or area chart uses a logarithmic scale on either the x or y axis. Using log scales with these mark types can produce misleading visuals and should be avoided.',
 'rect_without_d_d': 'A violation occurs if a rect mark is used and either the x or y channel is continuous. Rect marks require both axes to be discrete to represent a meaningful tiled layout.',
 'same_field_x_and_y': 'A violation occurs when the same field is assigned to both the x and y channels of a single mark. This redundancy creates a chart that is either meaningless or visually confusing.',
 'bar_tick_area_line_without_continuous_x_y': 'A violation occurs when a chart uses a bar, tick, area, or line mark but neither the x nor y channel is continuous. These marks depend on at least one continuous axis to effectively display measurements or trends.',
 'no_stack_with_bar_area_discrete_color': 'A violation occurs when a bar or area chart uses a discrete or binned color channel but does not use stacking. Stacking is required to accurately represent grouped values in this context.',
 'stack_without_discrete_color_or_detail': 'A violation occurs when stacking is enabled on a mark, but neither a discrete/binned color channel nor a detail channel is used. Stacking requires at least one of these to define how data should be grouped.',
 'stack_discrete': 'A violation occurs when stacking is applied to a channel that is discrete or binned. Stacking must only be applied to continuous channels to ensure correct rendering of data aggregation.',
 'encoding_field': 'Triggers when an encoding explicitly uses a field (i.e., `encoding.field` is defined). This suggests a preference to reduce the number of encodings that bind directly to data fields.',
 'same_field': 'Triggers when the same field is used exactly twice as an encoding for the same mark. This indicates a preference to avoid duplicating the same data field in multiple channels for a single mark.',
 'same_field_grt3': 'Triggers when the same field is used three or more times as an encoding for the same mark. This indicates a stronger penalty for repeatedly using the same field excessively.',
 'number_categorical': 'Triggers when a field of type `number` is encoded with a categorical scale type. This reflects a preference against treating numeric data as categorical.',
 'only_discrete': 'Triggers when a mark has no continuous encodings  all its channels are discrete or binned.',
 'multi_non_pos': 'Triggers when a single mark uses more than one non-positional channel (e.g., color, size, shape).',
 'non_pos_used_before_pos': 'Triggers when a non-positional channel is used in a mark but neither `x` nor `y` positional channels are present.',
 'cross_zero': 'Triggers when the data range for a field spans both negative and positive values, the encoding for that field is present, and the scale has `zero` set to `true`. This indicates a preference against forcing zero as a baseline in such cases.',
 'only_y': 'Triggers when a mark has an encoding for `y` but no encoding for `x`.',
 'high_cardinality_ordinal': 'Triggers when a field encoded with an ordinal scale has cardinality greater than 30.',
 'high_cardinality_categorical_grt10': 'Triggers when a field encoded with a categorical scale has cardinality greater than 10.',
 'high_cardinality_shape': 'Triggers when the shape channel is encoded with a field having cardinality greater than 8.',
 'high_cardinality_size': 'Triggers when the size channel is present, and the `x` or `y` positional encoding is continuous and has cardinality greater than 100.',
 'horizontal_scrolling_x': 'Triggers when the x-channel is discrete or binned and has cardinality greater than 50.',
 'log_scale': 'Triggers when an encoding uses a log scale type.',
 'ordinal_scale': 'Triggers when an encoding uses an ordinal scale type.',
 'categorical_scale': 'Triggers when an encoding uses a categorical scale type.',
 'c_c_line': 'Triggers when both x and y are continuous and the mark type is `line`.',
 'c_c_area': 'Triggers when both x and y are continuous and the mark type is `area`.',
 'c_d_overlap_point': 'Triggers when the x/y relationship is continuous by discrete, overlap is detected, and the mark type is `point`.',
 'c_d_overlap_bar': 'Triggers when the x/y relationship is continuous by discrete, overlap is detected, and the mark type is `bar`.',
 'c_d_overlap_line': 'Triggers when the x/y relationship is continuous by discrete, overlap is detected, and the mark type is `line`.',
 'c_d_overlap_area': 'Triggers when the x/y relationship is continuous by discrete, overlap is detected, and the mark type is `area`.',
 'c_d_no_overlap_point': 'Triggers when the x/y relationship is continuous by discrete, no overlap is detected, and the mark type is `point`.',
 'c_d_no_overlap_line': 'Triggers when the x/y relationship is continuous by discrete, no overlap is detected, and the mark type is `line`.',
 'c_d_no_overlap_area': 'Triggers when the x/y relationship is continuous by discrete, no overlap is detected, and the mark type is `area`.',
 'linear_color': 'Triggers when the color channel is used with a linear scale type.',
 'linear_size': 'Triggers when the size channel is used with a linear scale type.',
 'log_color': 'Triggers when the color channel is used with a log scale type.',
 'log_size': 'Triggers when the size channel is used with a log scale type.',
 'ordinal_x': 'Triggers when the x-channel is used with an ordinal scale type.',
 'ordinal_color': 'Triggers when the color channel is used with an ordinal scale type.',
 'ordinal_size': 'Triggers when the size channel is used with an ordinal scale type.',
 'ordinal_shape': 'Triggers when the shape channel is used with an ordinal scale type.',
 'categorical_color': 'Triggers when the color channel is used with a categorical scale type.',
 'polar_coordinate': 'Triggers when the view coordinates are set to `polar`.',
 'encoding': 'Triggers for each encoding entity present, indicating a preference to minimize the total number of encodings.',
 'c_c_point': 'Triggers when both x and y are continuous and the mark type is `point`.',
 'c_d_overlap_tick': 'Triggers when the x/y relationship is continuous by discrete, overlap is detected, and the mark type is `tick`.',
 'c_d_no_overlap_bar': 'Triggers when the x/y relationship is continuous by discrete, no overlap is detected, and the mark type is `bar`.',
 'd_d_point': 'Triggers when both x and y are discrete and the mark type is `point`.',
 'd_d_rect': 'Triggers when both x and y are discrete and the mark type is `rect`.',
 'log_x': 'Triggers when the x-channel uses a log scale type.',
 'log_y': 'Triggers when the y-channel uses a log scale type.',
 'ordinal_y': 'Triggers when the y-channel uses an ordinal scale type.'}
 \end{lstlisting}
 \label{app:NLprinciples}

\subsection{Visualization groups}

\paragraph{Encodings.}
Encodings define how data fields are translated into visual variables such as position, color, size, or shape. Rules in this group concern how data fields are mapped to these channels and whether the mapping is semantically appropriate for the data type and mark, detecting issues such as incorrect scale assignments, using non-positional channels without positional anchors, or reusing the same field redundantly across channels.

\paragraph{Marks.}
Marks correspond to the geometric primitives that represent data in a chart (e.g., bar, line, area, point, rect). Mark rules address whether the chosen primitive is suitable for the underlying data and encodings, flagging configurations where the mark type conflicts with axis types or scale choices—such as line/area charts with discrete axes or bars with fully continuous axes.

\paragraph{Stack.}
Stacking defines how multiple data values are visually accumulated along an axis in bar or area charts. Stacking rules evaluate whether this aggregation is correctly or incorrectly applied, detecting when stacking is required but missing (e.g., discrete color categories without stack) and when it is misused (e.g., stacking over discrete or binned fields without a grouping variable), ensuring that grouped values are represented accurately.

\paragraph{Scale.}
Scales describe how data values are transformed into visual space through mappings such as linear, logarithmic, ordinal, or categorical functions. Scale rules check whether the chosen scale type is compatible with the data and mark, identifying improper uses of log scales, inappropriate ordinal/categorical mappings, and misalignment between visual encodings and scale semantics.

\paragraph{Data.}
Data properties determine what the visualization receives as input—field types, cardinality, domain size, or coordinate transformations. Rules in this group address properties that impact readability and perception, such as excessive category counts, overly granular ordinal values, misuse of polar coordinates, or other data patterns that lead to cluttered or difficult charts.

 \label{app:VizGroups}

\end{document}